# What's the relationship between CNNs and communication systems?

Hao Ge | Xiaoguang Tu | Yanxiang Gong | Mei Xie* | Zheng Ma

***Abstract***—The interpretability of Convolutional Neural Networks (CNNs) is an important topic in the field of computer vision. In recent years, works in this field generally adopt a mature model to reveal the internal mechanism of CNNs, helping to understand CNNs thoroughly. In this paper, we argue the working mechanism of CNNs can be revealed through a totally different interpretation, by comparing the communication systems and CNNs. This paper successfully obtained the corresponding relationship between the modules of the two, and verified the rationality of the corresponding relationship with experiments. Finally, through the analysis of some cutting-edge research on neural networks, we find the inherent relation between these two tasks can be of help in explaining these researches reasonably, as well as helping us discover the correct research direction of neural networks.

**Index Terms**— Deep learning, adversarial example, interpretability

## 1. Introduction

Convolutional neural networks(CNNs) have demonstrated their superiority in many fields. However, the interpretability of CNNs is still in its infancy, which is the reason why we call them black boxes. Recently, an increasing number of researchers attempt to make this black box transparent. The methods can be summarized into the following three categories:

1.Network visualization [1],[5],[8], which is the most well-known method in the field of interpretability uses feature visualization methods to visualize what the neurons learned. These visualization methods give people a direct impression of the internal mechanism of neural networks.

2.Methods by semantic feature matching [3],[6], which aims to analyze the importance of different semantics in the processing of image classification, thus help people further understanding the neural networks from the perspective of features.

3.Methods by using mature models to simulate or explain neural networks [4],[9],[10], which is also the most mainstream research direction in the field of interpretability in recent years.

Despite the above-mentioned works, the issue of the interpretability of CNNs is still far from solved. Compared with the working mechanism of other filed that have been well studied, such as communication systems and circuit design, works in the field of CNNs lack theoretical foundation, modularity and industry standards. Therefore, the research of interpretability still has a long way to go.

In this paper, we adopt the method of the third category to interpret the CNNs. The benefit of this method is that there is relevant theoretical basis of particular mature models, which can be used to explain the CNNs to help us understand CNNs from a new perspective. We propose to compare the CNNs with a well studied model, the communication system, giving a new interpretation of CNN working mechanism. In the process of studying the adversarial examples [11], we accidently found that when the same stable distributed noises [7] are added to both the input images in CNNs and the channels in communication system, the classification accuracy decline curve and the bit error rate curve are very similar, as shown in figure 1 and figure 3. Therefore, we believe the CNNs and communication systems have at least some similarity when encountering noise interference. In addition, we propose a relation evaluation model between the modules of communication system and CNN as shown in Figure 2. We find this correspondence can help us using knowledge in the communication field to understand the latest research results in the field of neural networks. The details can be seen in section "Analysis".

## 2. Related Work

In this section, we review two tasks that closely relate to our model, the stable distribution [7] and the influence of such noise in the communication system [2].

### 2.1 Stable Distribution

Stable distributions [7] are a rich class of probability distributions that allow skewness and heavy tails and have many intriguing mathematical properties. In the early years, stable distributions were difficult to be practically used because lacking closed formulas for densities and distribution functions. However, with the development of computer technology, there are reliable computer programs to compute them now. So stable distribution have been widely used in many fields such as finance and communication.

By changing its α parameter, the stable distribution can be changed to many well-known distribution forms(eg: α=1: Cauchy distributions, α=2: Gaussian distributions, α=0.5: Levy distributions). This property makes stable distributions useful for data modeling in many scenarios. In the field of computer vision, there are two main types of noise, salt-pepper noise and Gaussian noise. These two types of noises can be generated by stable distributions with α equal to 0.9 and 2, respectively.

## 2.2 Communication System

Knowledge of communication systems is a pre-understanding of this paper. Therefore, we first briefly introduce communication systems. As can be seen in Figure 1, the communication system is composed of five modules, i.e., source coding, modulation, communication channel, demodulation and decoding. Source coding aims to encapsulate the original data into frames one by one according to the communication protocol. Each frame is an independent transmission unit, which includes the header and data. Modulation aims to first convert the digital signal into the low-frequency analog signal, and then add a carrier to it, so that the signals can be transmitted in a specified band. The communication channel involves three parts, transmit power, transmission medium and noise. The transmission medium determines the channel bandwidth, the ratio of transmit power and noise determines the signal-to-noise ratio(SNR). These two modules jointly determine the size of the channel capacity according to the Shannon formula: $C = W \log_2(1 + S/N)$. The demodulation and decoding are the inverse process of modulation and source coding, respectively.

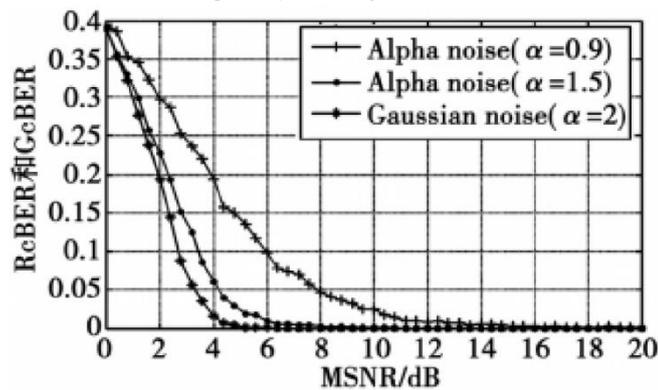

Fig.1: The influence of stable distributed noise on the bit error rate in the communication system.

In [2], the authors studied the influence of stable distributed noise on the communication system, and plotted the curve of the bit error rate as a function of the signal-to-noise ratio(SNR), as shown in Figure 1. The abscissa represents the SNR, and the ordinate represents the bit error rate. According to their observation, it's easy to draw the conclusion: when the remaining parameters of the alpha stable noise are the same, the smaller the alpha, the greater the bit error rate, especially when the SNR is relatively small.

# 3. Our Work

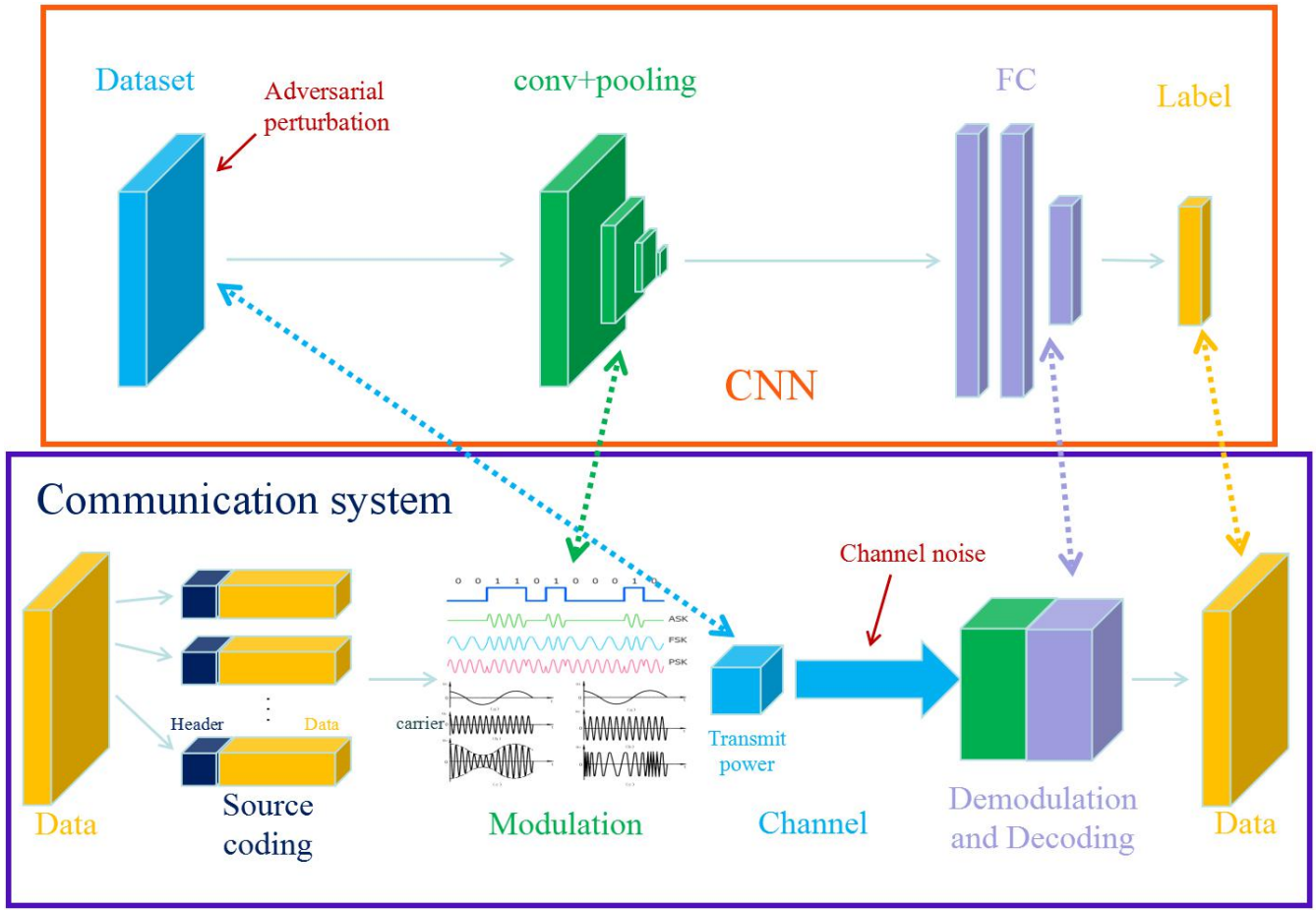

Fig.2: The correspondence between communication system and CNN.

We argue that there are many similarities between communication systems and CNNs. In Figure 2, we have marked their similar modules with the same color. We will explain them one by one following the structure of CNNs.

In the proposed relation evaluation model, the input of CNN can be viewed as the transmission power in the communication system. The larger number of the inputs of CNN indicate the more information of the dataset, meaning that the transmission power in the corresponding communication system is also larger. The adversarial disturbance in CNN can be viewed as the channel noise in the communication system. These two correspondences are the basis of this paper. The experimental section of this paper verifies the correctness of these correspondences.

The structure of the convolutional and pooling layers in CNN corresponds to the modulation and demodulation process in the communication system. In CNN, the role of the convolutional and pooling layer is to extract features, the amount of information used for classification in FC layer is carried on these features. In the communication system, the carrier in the modulation process is responsible for carrying information, which has the same function as the features in CNN. Therefore, CNN is similar to a communication system that divides information into many parts and uses different carrier frequencies to transmit information. The detailed explanation of this part can refer to the "Analysis" section of this paper.

The fully connected layer corresponds to the decoding process in the communication system. Both of them are used to summarize the information and give the final result. The label in cnn corresponds to the data in the communication system, as both of them are expected as accurate as possible.

There is a problem here that needs to be explained: as can be seen from Figure 2, the communication systems starts with "data", which corresponding to the "label" in CNNs. So the complete communication systems can be simulated to a "special" CNN: the labels are deconvoluted to generate the images, and the generated images are subsequently relabeled by convolutional, pooling and FC layers. In this perspective, the traditional CNNs are more similar to the second half of the communication systems(channel, demodulation and decoding). But we find it is more convenient to use modulation to explain the relationship between the carriers and features than to use demodulation. To this end, we use the structure as shown in Figure 2 to evaluate the relationship between the two tasks.

## 4. Experiment

Inspired by [2], we tested the decline in the classification accuracy of the classifier after adding stable distributed noises to the input images. For the purpose of reproducibility, we used the tf_flowers database [13] which can be easily downloaded with tensorflow to complete our experiments. The database is a subset of "oxford flowers", which contains 3670 images from 5 categories.

The horizontal axis in figure 3 represents the SNR of the input images, while the vertical axis is the difference in the classification accuracy of the classifier before and after the noises are added. The SNR can be calculated by formula (1).

$$SNR(dB) = 10\log_{10}(\frac{P_{Signal}}{P_{Noise}}) = 10\log_{10}(\frac{\Sigma_{i=1}^{3\times W\times L}(P_{ori})^2}{\Sigma_{i=1}^{3\times W\times L}(P_{ori} - P_{attacked})^2}) \quad (1)$$

W and L indicates the width and length of the image, $P_{ori}$ indicates the original pixel value, $P_{attacked}$ indicates the pixel value disturbed by noise.

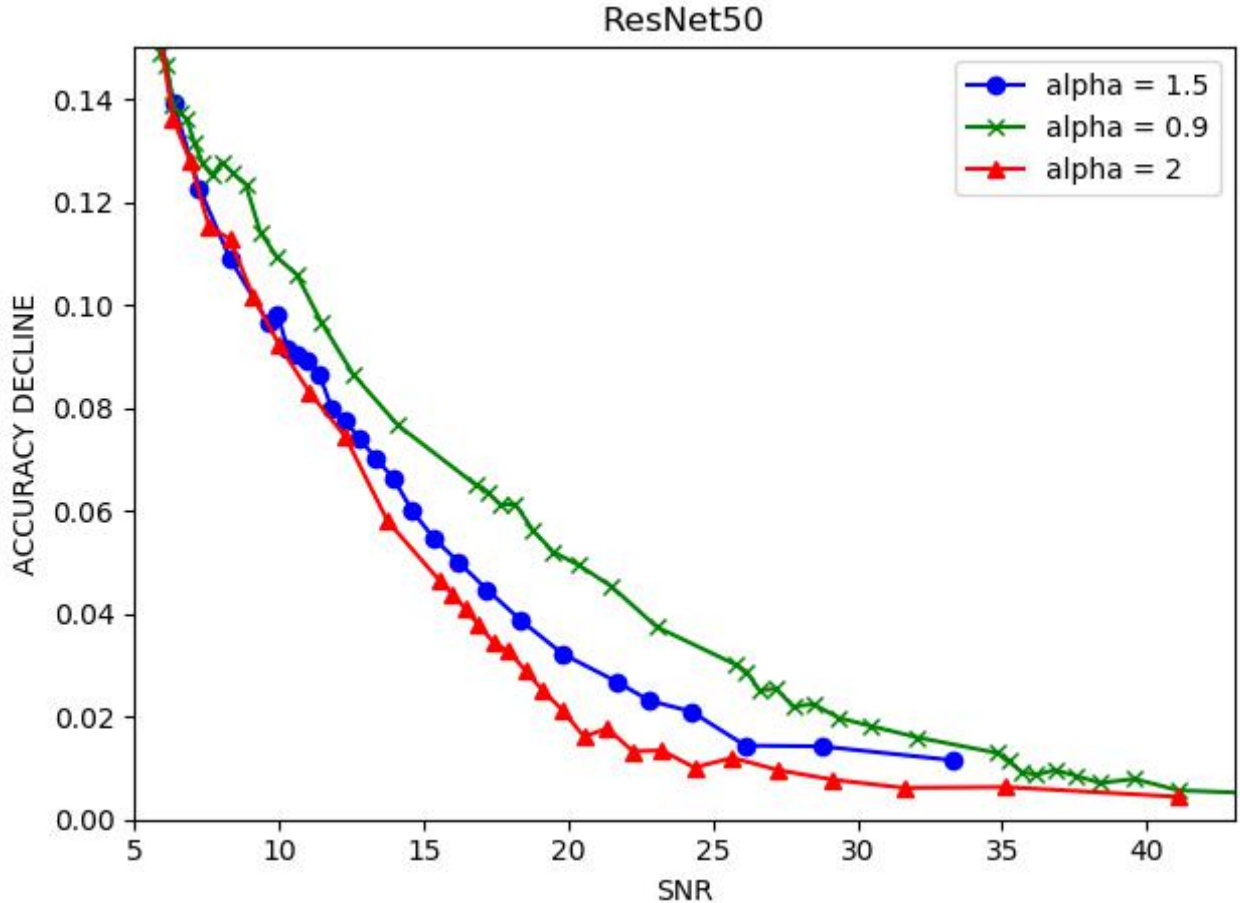

Fig.3: The performance of the stable distributed noises in CNN.

Comparing figure 1 and figure 3, we can find that the curves have same similarity. Based on this similarity, we try to describe the neural network with a communication system, so we got the corresponding way shown in Figure 2. We find that this correspondence can help us to understand some of the recent cutting-edge researches on neural networks. We will analyze them one by one in the next section.

In addition, we also test the improved version of salt-pepper noise($α$=0.9) by enlarging the noise points of salt-pepper noises to larger regions with specific shape(with same SNR), in order to test whether the more concentrated the noise information, the greater the impact on the classification accuracy of the classifier. We tested the noise of square, triangle and rhombus, respectively.

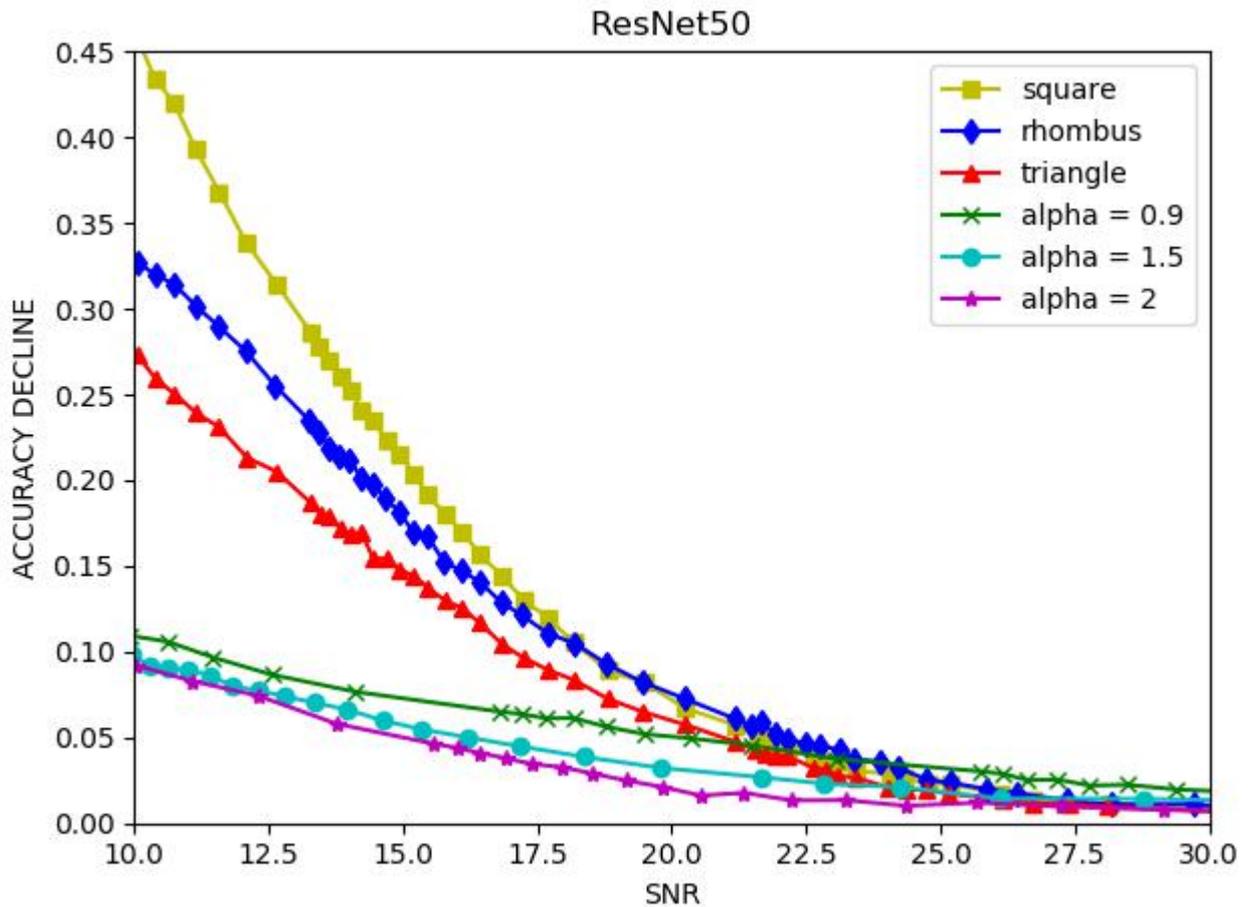

Fig.4: The performance of more concentrated noises.

As shown in Figure 4, it is obvious that these three kinds of noises that further concentrate the energy of the noises can bring a higher bit error rate than the salt-pepper noise($\alpha$=0.9). This result is consistent with our previous experimental results, because as the alpha decreases, the energy of the noise is also gradually concentrated. This result is also consistent with the performance of communication system, as the bit error rate of the communication system will also increase, when the impulse noises are concentrated in a small time window. By the way, all the above discussions are based on the same SNR.

## 5. Analysis

We argue if the two fields are related to each other, the emerging field can often learn from the other field, and this is the motivation of our method. For communication system, as a nearly century developed technology, it contains a large number of previous research results, including information theory, communication protocols, OSI layered models, and so on. We combine some of the latest research results in neural networks to explain why communication systems can help us understand neural networks and what directional guidance we can get from communication systems.

We first introduce the concept of several proper nouns. **Adversarial examples:** Well-designed input samples which can easily fool CNNs but imperceptible to human. **Robust learning:** A training process that learns features of real and adversarial samples simultaneously. **Standard learning:** A training process that only learns features of real samples.

### 5.1 Adversarial Examples Are Not Bugs, They Are Features

In [14], the author believes that the reason why the adversarial examples can affect the classification result of the classifiers is that the classifiers use a large number of non-robust features in the classification process, which are difficult to be observed by humans while truely existing in the real-world datasets, so when the adversarial perturbations change these non-robust features, the classification results of humans and classifiers will be different. They performed the following experiments on this view: they changed the labels of the adversarial examples to the results given by a classifier, and used these adversarial examples to train a new classifier, and then they found that the retrained classifier can successfully classify real images to their real labels. This experiment shows that adversarial perturbations really can bring some features of real images to adversarial examples.

This paper is very easy to understand from the perspective of communication coding. We know that in communication coding, all digital signals must be converted into analog signals, which are modulated by a carrier of a certain frequency (carrier frequency) to transmit information. The most effective interference method for this analog signal is to transmit noises of the same frequency. As shown in Figure 2, the channel noise in communication system is equivalent to the adversarial perturbation in image classification. Therefore, finding an effective adversarial perturbation is similar to finding the carrier frequency of a signal source in signal interference. As we know that the carrier frequency used by each signal source is different, the carrier frequency can naturally be called the characteristics of each signal source. Similarly, the adversarial examples also find the features that the classifier depends on, therefore, they do contain some of the features exist in the real images of the target category, but these features are not easily observed by humans, so the adversarial examples should not be called "bug", it can help us study what features the classification neural network relies on.

### 5.2 Adversarial Robust Generalization Requires More Data

In [15], the researchers show that the sample complexity of robust learning can be significantly larger than that of "standard" learning. This gap is information theoretic and independent of model selection and training method. So they postulate that the difficulty of training robust classifiers stems, at least partially, from this inherently larger sample complexity.

The point of this article can also be well understood from the perspective of communication coding. As mentioned earlier, the adversarial disturbance is similar to the noise at the same frequency as the carrier in the communication system, and this kind of noise will cause bit errors rates much higher than Gaussian white noise (under the premise of the same signal-to-noise ratio), so it is also very difficult to deal with this kind of noise in the communication field. The simplest way is to increase the transmission power, which makes the transmit power much larger than the noise power. As shown in Figure 2, the transmission power is equivalent to the amount of data in the image classification, so the view point of this paper can also well prove the correctness of our equivalent model.

### 5.3 Robustness May Be at Odds with Accuracy

In [16], the researchers show that training robust models may not only be more resource-consuming, but also lead to a reduction of standard accuracy. And they find that the features learned by robust models tend to align better with human perception.

As mentioned above, the features used by the classifier are similar to the carrier frequency used by the transmission source. We know that the classifiers utilize multiple features of the image to make classifications such as color, texture, material, object, etc. , so it is similar to the communication system using multiple carrier frequencies to transmit a part of information respectively, and finally collecting and analyzing at the receiving end. According to the experimental results of [3], the classifiers trained in Imagenet mainly rely on local features rather than global features. Therefore, adversarial perturbations are more inclined to interfere with these local features so that interfere with these classifiers to the greatest extent, then the robust model have to discard a part of the severely disturbed local features in order to correctly classify real images and adversarial examples at the same time, so the classification performance of the real images will naturally reduce. The features learned by the robust model is consistent with human perception as humans tend to use global features instead of local features when classifying images, and the robust model also discards local features that are seriously disturbed. It can be seen that considering the different features separately using the idea of communication coding can help us better understand the working mechanism of CNNs.

**5.4 Dynamic Routing Between Capsule**

The structure of CapsNet [17] is similar to the "frame" in the communication protocol(as shown in figure 2). Both of them encapsulate and process a piece of information and send the information to the next step. In communication protocols, an important role of frames is to ensure the accuracy of information transmission through check bits (when there is noise in the communication channel, the check bits can help the receiver to determine whether the received frame has bit errors). Therefore, we believe that this capsule structure will play an important role in the security field of neural networks in the future. In this point, we have done some related work [12]. We proposed to use a two-stream structure to detect adversarial examples. The two-stream structure consists of a high-resolution network and a low-resolution network as its check bit.

## 6. Conclusion

In this paper, by adding stable distributed noise to the original image, we find that CNN and communication system have similar properties when facing stable distributed noise. That is, when the signal-to-noise ratio is constant, the smaller the alpha, the greater the impact on the classification accuracy, especially when the signal-to-noise ratio is small. Based on this observation, the correspondence relationship between the modules of CNN and communication system is constructed, and the correctness of this correspondence relationship is verified in conjunction with recent developments in neural networks. It is believed that the revealed relationship between these two tasks can help researchers to find the correct development direction of neural networks more easily based on the development of communication systems.